\pdfoutput=1

\documentclass[11pt]{article}

\usepackage[preprint]{acl}
\usepackage{xcolor} 
\usepackage{adjustbox}
\usepackage{times}
\usepackage{latexsym}
\usepackage{booktabs}
\usepackage{multirow}
\usepackage[T1]{fontenc}
\usepackage[utf8]{inputenc}
\usepackage{microtype}
\usepackage{inconsolata}
\usepackage{graphicx}
\usepackage{comment}

\title{Dynamic Label Name Refinement for Few-Shot Dialogue Intent Classification}

\author{
Gyutae Park$^{1}$,~Ingeol Baek$^{1}$,~ByeongJeong Kim$^{1}$,~Joongbo Shin$^{2}$,~Hwanhee Lee$^{1}$\textsuperscript{$\dagger$} \\
    $^{1}$Department of Artificial Intelligence, Chung-Ang University, $^{2}$LG AI Research\\
    \texttt{\{pkt0401, ingeolbaek, michael97k, hwanheelee\}@cau.ac.kr}, \texttt{jb.shin@lgresearch.ai}
}

\begin{document}
\maketitle
\renewcommand*{\thefootnote}{\arabic{footnote}}

\begin{abstract}

Dialogue intent classification aims to identify the underlying purpose or intent of a user's input in a conversation. 
Current intent classification systems encounter considerable challenges, primarily due to the vast number of possible intents and the significant semantic overlap among similar intent classes.
In this paper, we propose a novel approach to few-shot dialogue intent classification through in-context learning, incorporating dynamic label refinement to address these challenges.
Our method retrieves relevant examples for a test input from the training set and leverages a large language model to dynamically refine intent labels based on semantic understanding, ensuring that intents are clearly distinguishable from one another.
Experimental results demonstrate that our approach effectively resolves confusion between semantically similar intents, resulting in significantly enhanced performance across multiple datasets compared to baselines.
We also show that our method generates more interpretable intent labels, and has a better semantic coherence in capturing underlying user intents compared to baselines.
\end{abstract}

\section{Introduction}
Dialogue intent classification identifies the underlying intent or purpose of a user's input in a conversation. It is a key component of task-oriented dialogue systems~\cite{degand2020introduction}, enabling accurate understanding of user utterances and generation of appropriate responses. However, current intent classification systems face challenges, particularly in managing a large number of intent classes and resolving semantic ambiguity between similar intents~\cite{sung-etal-2023-pre,cho2024zero,lu2024mitigatingboundaryambiguityinherent}.
Recent work explores few-shot learning approaches, including retrieval-augmented methods~\cite{milios2023incontextlearningtextclassification,gao2024ambiguityawareincontextlearninglarge,Abdullahi_2024} and prompt-based techniques~\cite{liang-etal-2023-breaking, parikh2023exploringzerofewshottechniques, zhang2024revisitfewshotintentclassification,rodriguez2024intentgptfewshotintentdiscovery}, which enable models to learn from limited examples per intent.
While retrieval-augmented methods effectively narrow down candidate intents by retrieving examples similar to the input query, these methods also introduce a critical challenge: the retrieved examples often show significant semantic overlap across different intent categories.

\begin{figure}[t] 
\centering
\includegraphics[width=0.9\linewidth]{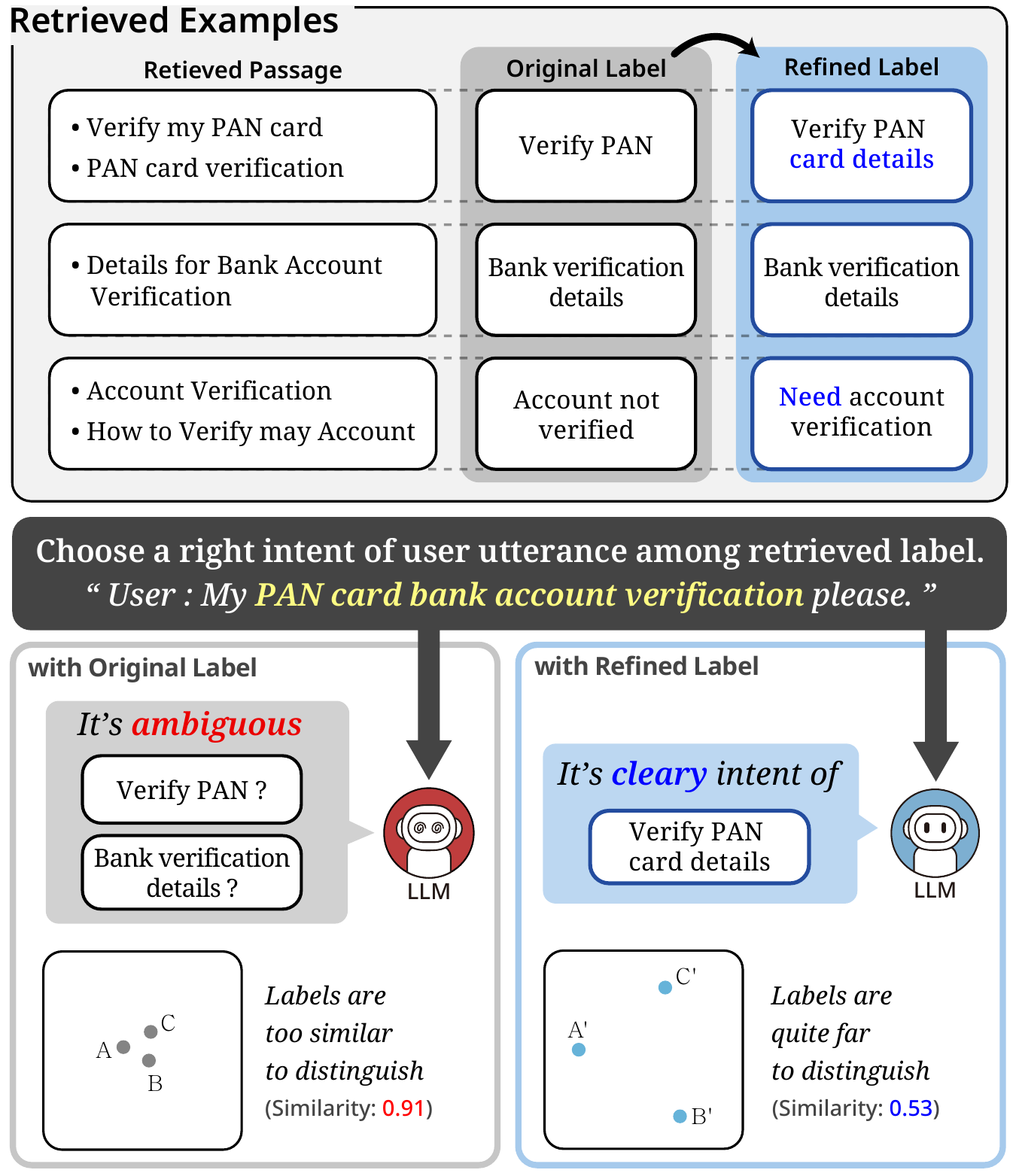}
\caption{An example illustrating how ambiguous and similar label names can confuse the model, while refined label names enable clearer decision-making.}
\label{fig:ds_len}
\vspace{-7mm}
\end{figure}

As shown in Figure ~\ref{fig:ds_len}, even with just three similar intents (`{Verify PAN}', `{Bank verification details}', and `{Account not verified}'), 
the model struggles to make accurate predictions due to their semantic similarity, as indicated by the cosine similarity score of 0.91 at the embedding space. 
We observe that this high semantic similarity between intent labels makes it challenging for models to distinguish between different intents accurately.


We find that these issues can be mitigated by refining the label names to forms that more distinctly differentiate them from other labels. 
As illustrated in Figure~\ref{fig:ds_len}, by mapping the label `{Verify PAN}' to a more descriptive form, such as `{Verify PAN card details}', it becomes easier to differentiate it from general bank verification intents.
This refinement establishes clearer semantic boundaries between intent categories, resulting in more accurate classifications.

In this paper, we present a novel approach that combines dynamic label refinement with similarity-based example selection. Our method involves retrieving semantically similar examples and dynamically refining their intent labels to create more meaningful distinctions between related intents. Through extensive experiments across various model scales and diverse datasets, we achieve significant improvements.
Our analysis shows that the improvements are particularly pronounced in datasets with high semantic overlap between intents, with accuracy gains ranging from 2.07\% to 7.51\% across different model scales.

\section{Method}
\begin{figure}
    \centering
    \includegraphics[width=0.95\linewidth]{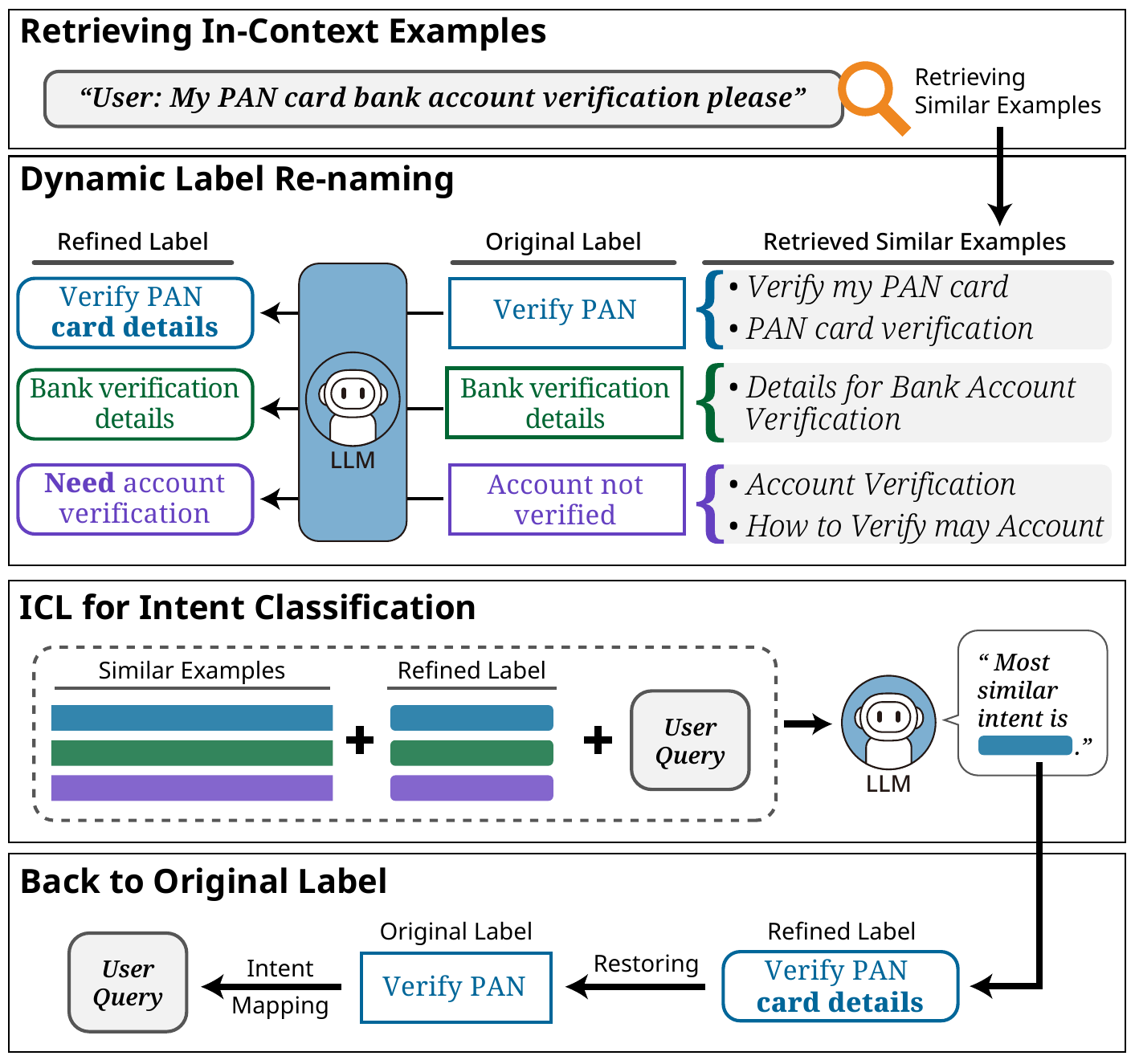}
    \caption{Overall flow of the proposed dynamic label name refinement method for intent classification.}
    \label{fig:enter-label}
    \vspace{-5mm}
\end{figure}
Following the recent works in dialogue intent classification ~\cite{milios2023incontextlearningtextclassification,chen2024retrievalstyleincontextlearningfewshot,chandra2024sizedoesntfitall}, our approach leverages retrieval-based in-context learning (ICL) with large language models (LLMs), which has demonstrated effectiveness in tasks involving large label spaces. This approach allows models to dynamically leverage relevant few-shot examples for prediction from the training set. 

We introduce a retrieval ICL method for intent classification with dynamic label re-naming, which comprises three steps: (1) retrieving semantically similar examples, (2) refining intent labels using an LLM to generate more descriptive labels, and (3) conducting the final classification with these refined examples.

\subsection{Retrieving In-Context Examples}
We start by retrieving relevant examples in the dataset for each input query. To retrieve such semantically similar examples, we use a pre-trained SentenceTransformer model~\cite{reimers2019sentence}. For each input query, we retrieve the top-20 most similar examples. These retrieved examples are then grouped by their original intents to provide comprehensive context for the subsequent steps.

\subsection{Dynamic Label Re-naming}
Our proposed dynamic label refinement process combines retrieval-based example selection with label generation guided by an LLM, as shown in Figure~\ref{fig:enter-label}. 
For each of the specified intent groups, we design tailored instructions for the LLM to analyze these groups and refine the labels accordingly. (see Appendix~\ref{appendix:prompt_details} for full prompt)
Specifically, we design intent label refinement as a process where the model evaluates whether to retain the original label or propose an enhanced version while preserving the domain-specific semantics. During this process, the model assesses the semantic relationship between the label and its associated examples, deciding either to maintain the original intent name or to generate a more descriptive alternative.
For instance, as shown in Figure~\ref{fig:enter-label}, when analyzing examples like ``\textit{Verify my PAN card}'' and ``\textit{Pan card verification}'', the model recognizes that the original label `\texttt{verify\_pan}' could be more descriptive and refines it to `\texttt{verify\_pan\_card\_details}' to better capture the specific verification intent.

\begin{table*}[!ht]
\centering
\resizebox{1.0\textwidth}{!}{
\footnotesize
\setlength{\tabcolsep}{3pt}
\renewcommand{\arraystretch}{1.2}
\begin{tabular}{l|ccc|ccc|ccc}
\hline
\multirow{2}{*}{Dataset} & \multicolumn{3}{c|}{Llama3-8b-inst} & \multicolumn{3}{c|}{Qwen2.5-7b-inst} & \multicolumn{3}{c}{Qwen2.5-1.5b-inst} \\ \cline{2-10}
& \textbf{Raw} & \textbf{CoT} & \textbf{Refined} & \textbf{Raw} & \textbf{CoT} & \textbf{Refined} & \textbf{Raw} & \textbf{CoT} & \textbf{Refined} \\ \hline
HWU64 & 88.10 & 87.17 {\scriptsize\color{blue}(-0.93)} & \textbf{89.03} {\scriptsize\color{red}(+0.93)} & 87.08 & 86.71 {\scriptsize\color{blue}(-0.37)} & \textbf{88.38} {\scriptsize\color{red}(+1.30)} & 78.90 & 78.72 {\scriptsize\color{blue}(-0.18)} & \textbf{80.76} {\scriptsize\color{red}(+1.86)} \\
BANKING77 & 85.88 & 85.48 {\scriptsize\color{blue}(-0.40)} & \textbf{87.95} {\scriptsize\color{red}(+2.07)} & 85.68 & 87.05 {\scriptsize\color{red}(+1.37)} & \textbf{87.30} {\scriptsize\color{red}(+1.62)} & 73.31 & 72.63 {\scriptsize\color{blue}(-0.68)} & \textbf{77.34} {\scriptsize\color{red}(+4.03)} \\
CLINC150 & 95.03 & 95.13 {\scriptsize\color{red}(+0.10)} & \textbf{95.51} {\scriptsize\color{red}(+0.48)} & 95.36 & 95.42 {\scriptsize\color{red}(+0.06)} & \textbf{95.58} {\scriptsize\color{red}(+0.22)} & 82.29 & 81.90 {\scriptsize\color{blue}(-0.39)} & \textbf{84.02} {\scriptsize\color{red}(+1.73)} \\ \hline
CUREKART & 89.76 & 89.76 {\scriptsize\color{blue}(+0.00)} & \textbf{91.94} {\scriptsize\color{red}(+2.18)} & 90.02 & 89.98 {\scriptsize\color{blue}(-0.04)} & \textbf{90.40} {\scriptsize\color{red}(+0.38)} & 82.78 & 82.35 {\scriptsize\color{blue}(-0.43)} & \textbf{84.10} {\scriptsize\color{red}(+1.32)} \\
POWERPLAY11 & 70.87 & 67.00 {\scriptsize\color{blue}(-3.87)} & \textbf{76.10} {\scriptsize\color{red}(+5.23)} & 71.20 & 70.06 {\scriptsize\color{blue}(-1.14)} & \textbf{74.76} {\scriptsize\color{red}(+3.56)} & 60.84 & 58.83 {\scriptsize\color{blue}(-2.01)} & \textbf{63.10} {\scriptsize\color{red}(+2.26)} \\
SOFMATTRESS & 82.61 & 81.10 {\scriptsize\color{blue}(-1.51)} & \textbf{85.40} {\scriptsize\color{red}(+2.79)} & 83.79 & 84.46 {\scriptsize\color{red}(+0.67)} & \textbf{87.40} {\scriptsize\color{red}(+3.61)} & 73.12 & 73.70 {\scriptsize\color{red}(+0.58)} & \textbf{80.63} {\scriptsize\color{red}(+7.51)} \\ \hline
\end{tabular}
}
\caption{Performance comparison of our dynamic label refinement approach across different models. \textbf{Raw} represents the baseline performance using original intent labels, \textbf{CoT} shows results with chain-of-thought prompting, and \textbf{Refined} shows the results after applying our dynamic label refinement method.}
\label{tab:main}
\vspace{-3mm}
\end{table*}



\subsection{ICL for Intent Classification}
After obtaining the new labels for each sample, we leverage ICL with the refined labels and examples for final classification. This two-step process where the same LLM both refines the labels and makes the final classification decision helps ensure consistency between the refined semantic understanding and the ultimate intent prediction. Consequently, the overall process involves constructing a prompt that includes:
1) The retrieved examples with their refined intent labels
2) The test query requiring classification
3) Clear instructions for the model to select the most appropriate intent

\section{Experiment}
\subsection{Experimental Setup}
\paragraph{Datasets}

We evaluate our method on two groups of datasets: DialoGLUE benchmark datasets \cite{mehri2020dialogluenaturallanguageunderstanding} (BANKING77, HWU64, CLINC150) and HINT3 datasets \cite{arora-etal-2020-hint3} (CUREKART, POWERPLAY11, SOFMATTRESS). For DialoGLUE datasets, we follow the standard 10-shot setting where each intent has only 10 examples for training. For HINT3 datasets, we exclude out-of-scope queries (NO\_NODES\_DETECTED) from evaluation as they are irrelevant to our intent classification task. Further details about datasets are provided in Appendix~\ref{dataset details}.


\paragraph{Models}
We conduct experiments with three different sizes of LLMs to evaluate the effectiveness of our approach. We employ \textit{Llama3-8b-inst.}~\cite{dubey2024llama}, \textit{Qwen2.5-7b-inst.}~\cite{yang2024qwen2}, and \textit{Qwen2.5-1.5b-inst}. as our backbone models.

\paragraph{Baselines}
We compare our approach with the following baselines:
1) \textbf{Raw}: A standard in-context learning approach where we retrieve similar examples and perform intent classification using the original intent labels without any refinement.
2) \textbf{Chain-of-Thought (CoT)} \cite{wei2023chainofthoughtpromptingelicitsreasoning}: An enhanced prompting method that guides the model to break down the intent classification process into steps, analyzing the user query and retrieved examples before making a prediction. 

\paragraph{Implementation Details}
Following~\cite{milios2023incontextlearningtextclassification}, we order examples from least to most similar in the prompt, which demonstrated higher accuracy across our datasets. We use this retrieval-based in-context learning setup as our baseline, where the LLM directly performs classification using the original intent labels.
While a static label refinement might seem simpler, we opt for dynamic refinement as it enables context-specific label adjustments based on each test query and its retrieved examples. We also confirm this through the experiment in Appendix~\ref{dynamic vs static comparison}.

\subsection{Main Results}

Table~\ref{tab:main} presents a comprehensive analysis of our dynamic label refinement approach across different experimental setups. 
We structure the investigation around several critical dimensions to clearly demonstrate the effectiveness and implications of our method.

\paragraph{Semantic Disambiguation through Label Refinement}
We first confirm that while both Raw and CoT approaches show reasonable performance, they often struggle with semantically ambiguous intents. For example, CoT misclassifies ``\textit{Please keep delivery service to the pin code 7021}'' as `\texttt{modify\_address}' instead of `\texttt{check\_pincode}', and "\textit{Hey I didn't receive the ordered product its incomplete}" as `\texttt{refunds\_returns\_replacements}' instead of `\texttt{delay\_in\_parcel}'. 
These examples show how structured reasoning often struggles when similar intents have overlapping semantics.
Our refinement process tackles this issue by analyzing the semantic relationships between labels and examples, leading to consistent performance improvements over both Raw and CoT baselines across all datasets as shown in Table~\ref{tab:main}.
For example, in BANKING77, this approach achieves notable improvements of +2.07\%, +1.62\% and +4.03\% for Llama3-8b-inst., Qwen2.5-7b-inst., and Qwen2.5-1.5b-inst. respectively, demonstrating the effectiveness of semantic-aware label refinement. The improvements are particularly notable in datasets with complex domain-specific terminology, such as BANKING77, where the model effectively leverages existing semantic information in the labels.

\paragraph{Performance Across Model Scales}
Our experiments with different model sizes reveal several interesting patterns about the scalability of our approach. 
The larger models (Llama3-8b-inst. and Qwen2.5-7b-inst.) demonstrate robust baseline performance with moderate improvements (2.07\% and 1.62\% on BANKING77, 3.61\% and 3.56\% on SOFMATTRESS respectively). Notably, even the smaller Qwen2.5-1.5b-inst. Model achieves significant improvements, particularly on domain-specific datasets (+7.51\% on SOFMATTRESS, +4.03\% on BANKING77), suggesting that our approach effectively enhances performance regardless of model scale. These results demonstrate that our proposed method effectively enhances intent classification performance across various datasets and model architectures.

\subsection{Analysis}

\paragraph{Semantic Similarity Analysis}
To validate our hypothesis about semantic relationships between intent labels, we conduct an embedding-based similarity analysis comparing original and refined intent labels across all datasets. Specifically, we leverage the last hidden layer representations from our LLMs to capture the semantic characteristics of each intent label. For each intent, we extracte the final hidden state representation and computed pairwise cosine similarities between these representations within each label set.

\begin{table}[h!]
\centering
\small
\renewcommand{\arraystretch}{1.4}
\begin{tabular}{p{2.6cm}cc}
\hline
\textbf{Model} & \textbf{Original} & \textbf{Refined} \\
\hline
Llama3-8b-inst. & 0.86 & \textbf{0.74} \\
Qwen2.5-7b-inst. & 0.83 & \textbf{0.80} \\
Qwen2.5-1.5b-inst. & 0.95 & \textbf{0.91} \\
\hline
\end{tabular}
\caption{Average pairwise semantic similarity between original and refined intent labels across various model scales.}
\label{tab_similarity}
\vspace{-3mm}
\end{table}

We observe that refined labels consistently achieve lower average pairwise similarities than original labels across all datasets and model scales. 
With Llama3-8b-inst., the average similarity drops from 0.86 for original intent labels to 0.74 for refined labels. 
Qwen2.5-7b-inst. and Qwen2.5-1.5b-inst. follow a similar trend, with reductions from 0.83 to 0.80 and 0.95 to 0.91, respectively, although the decrease is less significant compared to Llama3-8b-inst.
This decrease in semantic overlap directly correlates with improved classification performance. For instance, on the BANKING77 dataset, Llama3-8b-inst. achieves a 2.07\% accuracy improvement along with a 0.1 reduction in label similarity. The performance gains become especially noticeable when the model creates more semantically distinct labels, indicating that reducing label overlap enables clearer distinctions between different intents. We also provide the detailed similarity analysis for each dataset in Appendix~\ref{Dataset-wise Similarity Analysis}.

\begin{table}[t]
\centering
\adjustbox{width=0.48\textwidth}{ 
\setlength{\tabcolsep}{3pt}
\renewcommand{\arraystretch}{1.1}
\begin{tabular}{l||cc|cc}
\hline
\multirow{2}{*}{Dataset} & \multicolumn{2}{c|}{Q2.5-7B} & \multicolumn{2}{c}{Q2.5-1.5B} \\ 
\cline{2-5}
& Q2.5-7B & Q2.5-1.5B  & Q2.5-1.5B & Q2.5-7B \\ 
\hline
HWU64 & \textbf{88.38} & 81.22  & 80.76 & \textbf{88.01} \\
BANKING77 & \textbf{87.30} & 80.81& 77.34 & \textbf{85.95} \\
CLINC150 & \textbf{95.58} & 87.31& 84.02 & \textbf{95.22} \\
CUREKART & \textbf{90.40} & 82.78 & 84.10 & \textbf{88.90} \\
POWERPLAY11 & \textbf{74.76} & 64.10 & 63.10 & \textbf{74.43} \\
SOFMATTRESS & \textbf{87.40} & 78.30 & 80.63 & \textbf{85.08} \\ 
\hline
\end{tabular}%
}
\caption{Performance comparison across different model combinations. The top row means the model used for re-naming, and the second row denotes the models used for classification. Models used: Qwen2.5-7B-inst. (Q2.5-7B) and Qwen2.5-1.5B-inst. (Q2.5-1.5B).} 
\vspace{-5mm}
\label{tab:model-combinations}
\end{table}
\paragraph{Model Combination}
To validate the effectiveness of labeling refinement itself of each model, we employ two separate LLMs for intent refinement and classification tasks. We experiment with various combinations of Qwen2.5-7B-inst. and Qwen2.5-1.5B-inst. As in Table~\ref{tab:model-combinations}, using a larger model for refinement followed by a smaller model for classification yields the best performance. For example, using Qwen2.5-7B-inst. for refinement and Qwen2.5-1.5B-inst. for answering achieves notable improvements: +3.31\% on CLINC150 and +4.10\% on POWERPLAY11 compared to single-model baselines.

Interestingly, even reverse combinations (small model refinement + large model classification) show improvements over the non-refinement baseline, though to a lesser extent. For instance, using Qwen2.5-1.5B-inst. for refinement and Qwen2.5-7B-inst. for answering still achieves improvements on the datasets. This suggests that our label refinement approach is robust across different model scales and configurations. 
\section{Conclusion}

We propose a dynamic label refinement method for few-shot dialogue intent classification that mitigates the issues of significant semantic overlap between intent labels. Using the retrieved examples, we refine labels via LLMs to create more semantically distinct intent categories. Experimental results demonstrate that our method consistently improves performance across multiple datasets for various models. We also confirme that our method reduces semantic similarities between intent labels, creating more distinct and interpretable categories.
\section*{Limitations}
While our approach demonstrates significant improvements in intent classification performance, it requires additional computational overhead compared to traditional methods. The need to run the model once for label refinement and once for classification - increases the computational cost per query. However, we believe this trade-off is justified by the substantial improvements in classification accuracy, particularly for semantically ambiguous intents.
\section*
{Acknowledgement}
This research was supported by Institute for Information \& Communications Technology Planning \& Evaluation (IITP) through the Korea government (MSIT) under Grant No. 2021-0-01341 (Artificial Intelligence Graduate School Program (Chung-Ang University)).



\bibliography{custom}
\clearpage
\appendix

\section{Dataset Details}
\label{dataset details}
We evaluate our approach using two different groups of datasets. The first group consists of DialoGLUE benchmark datasets (\textbf{BANKING77}: 77 intents focused on banking domain~\cite{casanueva-etal-2020-efficient}, \textbf{HWU64}: 64 intents spanning across 21 domains~\cite{casanueva-etal-2020-efficient}, \textbf{CLINC150}: 150 intents covering 10 domains~\cite{larson-etal-2019-evaluation}), which are widely used for evaluating task-oriented dialogue systems. For these datasets, we follow the standard 10-shot setting where each intent class has only 10 examples for training, reflecting real-world scenarios where collecting large amounts of labeled data is challenging.

The second group includes HINT3 datasets~\cite{arora-etal-2020-hint3} (\textbf{CUREKART}: 28 intents in fitness supplements retail domain, \textbf{POWERPLAY11}: 59 intents in online gaming domain, \textbf{SOFMATTRESS}: 21 intents in mattress products retail domain), which contain real user queries from live chatbots. We exclude the \texttt{NO\_NODES\_DETECTED} label from the test set as it represents out-of-scope queries irrelevant to our task.

\section{Dynamic vs Static Label Refinement}
\label{dynamic vs static comparison}
\begin{table}[h]
\renewcommand{\arraystretch}{1.2}
\resizebox{0.5\textwidth}{!}{
\footnotesize
\setlength{\tabcolsep}{3.5pt}
\centering
\begin{tabular}{l|ccc|ccc}
\toprule
\multirow{2}{*}{Dataset} & \multicolumn{3}{c|}{Llama3-8b-inst.} & \multicolumn{3}{c}{Qwen2.5-1.5b-inst.} \\
\cmidrule(lr){2-4} \cmidrule(lr){5-7}
& Baseline & Static & Dynamic & Baseline & Static & Dynamic \\
\midrule
HWU64 & 88.10 & 84.10 {\scriptsize\color{blue}(-4.00)} & \textbf{89.03} {\scriptsize\color{red}(+0.93)} & 78.90 & 74.90 {\scriptsize\color{blue}(-4.00)} & \textbf{80.76} {\scriptsize\color{red}(+1.86)} \\
BANKING77 & 85.88 & 81.65 {\scriptsize\color{blue}(-4.23)} & \textbf{87.95} {\scriptsize\color{red}(+2.07)} & 73.31 & 69.18 {\scriptsize\color{blue}(-4.13)} & \textbf{77.34} {\scriptsize\color{red}(+4.03)} \\
CLINC150 & 95.03 & 87.60 {\scriptsize\color{blue}(-7.43)} & \textbf{95.51} {\scriptsize\color{red}(+0.48)} & 82.29 & 83.42 {\scriptsize\color{red}(+1.13)} & \textbf{84.02} {\scriptsize\color{red}(+1.73)} \\
\midrule
CUREKART & 89.76 & 85.47 {\scriptsize\color{blue}(-4.29)} & \textbf{91.94} {\scriptsize\color{red}(+2.18)} & 82.78 & \textbf{86.27} {\scriptsize\color{red}(+3.49)} & 84.10 {\scriptsize\color{red}(+1.32)} \\
POWERPLAY11 & 70.87 & 68.61 {\scriptsize\color{blue}(-2.26)} & \textbf{76.10} {\scriptsize\color{red}(+5.23)} & 60.84 & 55.66 {\scriptsize\color{blue}(-5.18)} & \textbf{63.10} {\scriptsize\color{red}(+2.26)} \\
SOFMATTRESS & 82.61 & 83.00 {\scriptsize\color{red}(+0.39)} & \textbf{84.60} {\scriptsize\color{red}(+1.99)} & 73.12 & 73.91 {\scriptsize\color{red}(+0.79)} & \textbf{80.63} {\scriptsize\color{red}(+7.51)} \\
\bottomrule
\end{tabular}
}
\caption{Performance comparison of baseline, static and dynamic refinement approaches (\%). All changes are computed relative to baseline. Best results for each dataset are marked in \textbf{bold}, improvements are shown in \textcolor{red}{red}, and decreases are shown in \textcolor{blue}{blue}.}
\label{tab:performanc_comparison}
\end{table}

We conduct a detailed comparison between dynamic and static label refinement approaches to validate our method choice. In the static approach, intent labels are refined once before training using all training examples collectively, following the same refinement rules as our dynamic approach. While this might seem advantageous in terms of computational efficiency and consistency, our experimental results demonstrate that the dynamic refinement approach consistently outperforms the static baseline across all datasets and model architectures.

Using Llama3-8b-inst., the dynamic approach achieves +2.07\% improvement on BANKING77, while static refinement shows a -4.23\% decrease. Similarly on CLINC150, dynamic refinement yields +0.48\% improvement, compared to -7.43\% with static refinement. Even with smaller models like Qwen2.5-1.5b-inst., dynamic refinement shows consistent gains (+4.03\% on BANKING77) while static refinement often leads to performance degradation.

The superior performance of dynamic refinement can be attributed to several factors. Dynamic refinement considers the specific context of each test query and its retrieved examples, allowing for more nuanced label adjustments. By refining labels based on retrieved similar examples, the model can better capture the semantic relationships specific to the current query context. Additionally, the dynamic approach can adjust its refinement strategy based on the semantic similarity patterns observed in the retrieved examples, rather than using fixed refined labels.

\section{Impact of the Number of Retrieved Examples}

\begin{figure}[h]
\centering
\begin{minipage}{0.45\textwidth}
    \centering
    \includegraphics[width=\textwidth]{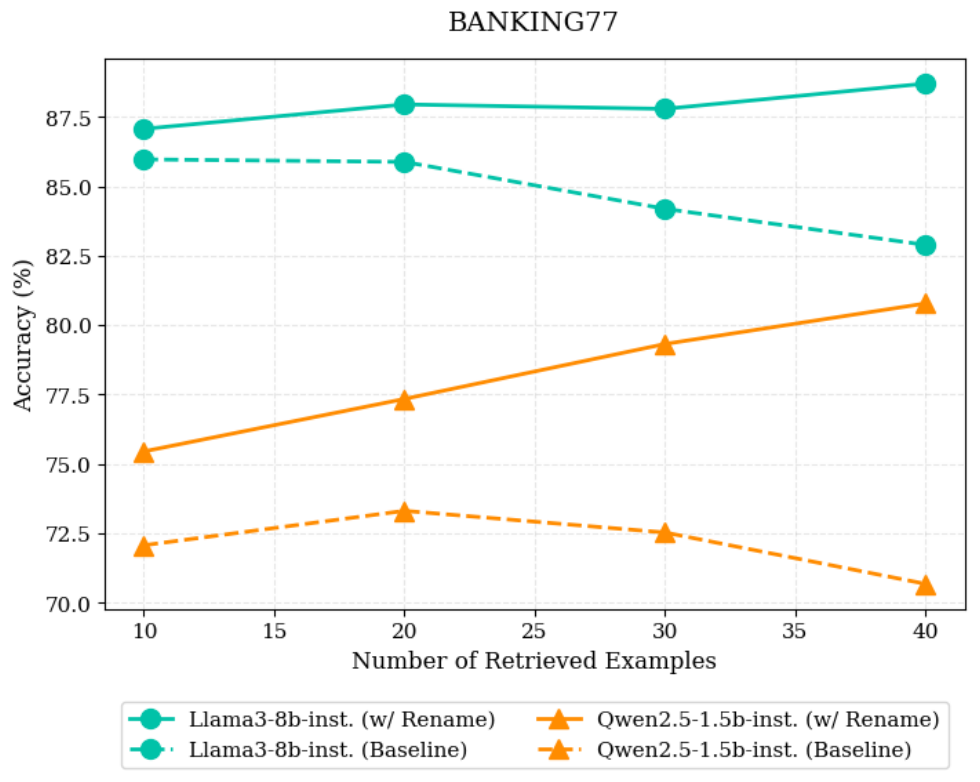}
    \caption*{BANKING77}
\end{minipage}

\vspace{0.2cm}

\begin{minipage}{0.45\textwidth}
    \centering
    \includegraphics[width=\textwidth]{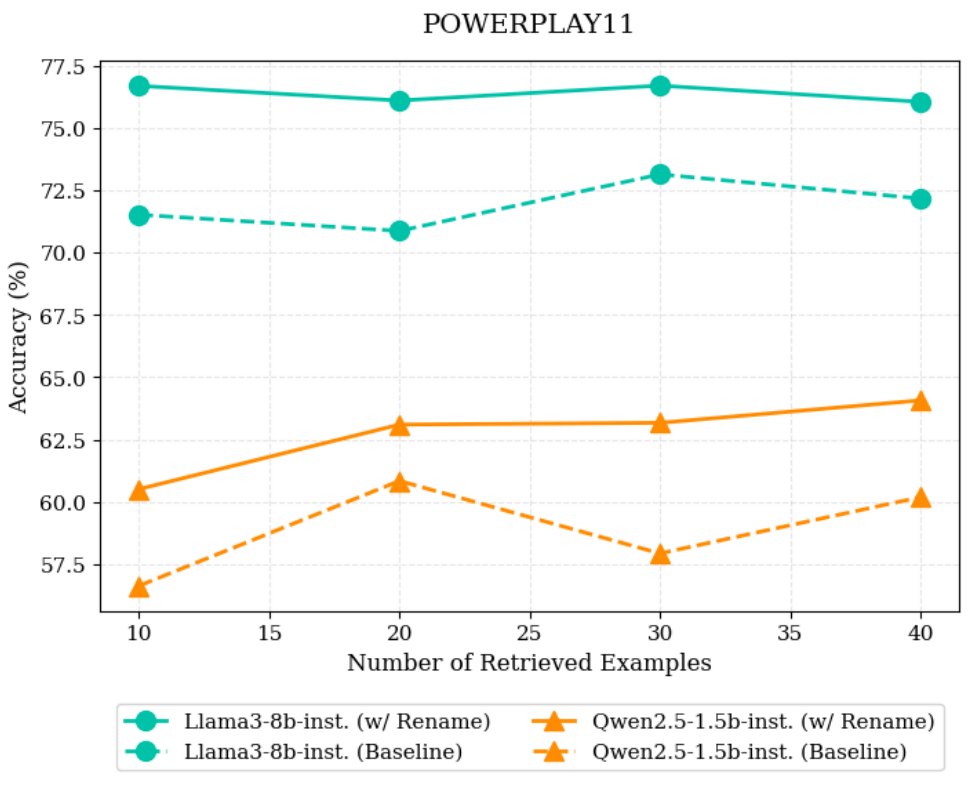}
    \caption*{POWERPLAY11}
\end{minipage}
\caption{Performance comparison with different numbers of retrieved examples. Solid lines represent performance with label refinement, while dashed lines represent baseline performance without refinement.}
\label{fig:retrieval_counts_impact}
\end{figure}

We conduct additional experiments to analyze the impact of the number of retrieved examples on model performance. Figure ~\ref{fig:retrieval_counts_impact} shows the performance comparison between different model configurations across varying numbers of retrieved examples (10, 20, 30, and 40) on BANKING77 and POWERPLAY11 datasets.

From these results, we observe several key patterns:

• The performance gap between the baseline and refined versions tends to be more pronounced with larger numbers of examples, particularly in BANKING77.

• The larger model (Llama3-8b-inst.) shows more stable performance across different example counts, while the smaller model (Qwen2.5-1.5b-inst.) shows greater variance in performance.

• POWERPLAY11 shows relatively consistent improvement patterns across different example counts, suggesting that the benefits of label refinement are robust across different dataset characteristics.

These findings suggest that our label refinement approach is effective across different numbers of retrieved examples.

\begin{table}[h]
\begin{tabular}{lcc}
\toprule
Dataset & Recall@20 & Avg. Intents \\
\midrule
HWU64 & 97.77 & 7.54 \\
BANKING77 & 98.93 & 7.04 \\
CLINC150 & 99.33 & 6.31 \\
CUREKART & 98.91 & 4.31 \\
SOFMATTRESS & 98.81 & 5.86 \\
POWERPLAY11 & 96.44 & 6.93 \\
\bottomrule
\end{tabular}
\caption{Retrieval analysis with 20 examples}
\end{table}

\section{Dataset-wise Similarity Analysis}
\label{Dataset-wise Similarity Analysis}
To provide a detailed view of semantic similarity patterns, we analyze the pairwise similarities between intent labels for each dataset. Table 8 shows the average pairwise similarity scores for both original and refined intent labels across different datasets and models.

\begin{table}[h]
\small
\renewcommand{\arraystretch}{1.2}
\setlength{\tabcolsep}{4pt}
\resizebox{0.5\textwidth}{!}{%
\centering
\begin{tabular}{lcccccc}
\hline
\multirow{2}{*}{Dataset} & \multicolumn{2}{c}{Llama3-8b-inst.} & \multicolumn{2}{c}{Qwen2.5-7b-inst.} & \multicolumn{2}{c}{Qwen2.5-1.5b-inst.} \\
\cmidrule(lr){2-3} \cmidrule(lr){4-5} \cmidrule(lr){6-7}
& Original & Refined & Original & Refined & Original & Refined \\
\hline
HWU64 & 0.835 & 0.721 & 0.772 & 0.760 & 0.948 & 0.910 \\
BANKING77 & 0.889 & 0.786 & 0.852 & 0.838 & 0.955 & 0.938 \\
CLINC150 & 0.834 & 0.705 & 0.780 & 0.764 & 0.946 & 0.901 \\
CUREKART & 0.881 & 0.802 & 0.800 & 0.833 & 0.956 & 0.925 \\
SOFMATTRESS & 0.849 & 0.761 & 0.752 & 0.748 & 0.943 & 0.911 \\
POWERPLAY11 & 0.844 & 0.688 & 0.838 & 0.812 & 0.945 & 0.903 \\
\hline
Mean & 0.856 & 0.744 & 0.799 & 0.793 & 0.949 & 0.915 \\
\hline
\end{tabular}
}
\caption{Detailed semantic similarity analysis across datasets and models. Values represent the average pairwise cosine similarity between all intent labels within each dataset. Lower values indicate more semantically distinct intent categories.}
\label{tab:detailed_similarity_results}
\end{table}

As shown in Table~\ref{tab:detailed_similarity_results}, the reduction in semantic similarity is consistent across all datasets for both models, though the magnitude of reduction varies. For Llama3-8b-inst., POWERPLAY11 shows the largest reduction in similarity (0.156), while CUREKART shows the smallest (0.079). Similarly, for Qwen2.5-1.5b-inst., CLINC150 shows a notable reduction (0.045) while BANKING77 shows a relatively smaller change (0.017). These variations might reflect differences in the initial intent label structures across datasets and the models' ability to refine them effectively.

\section{Semantic vs Generic Label Refinement}
\label{Semantic vs Generic Label Refinemetn}
\begin{table*}[h]
\small
\centering
\setlength{\tabcolsep}{4pt}
\renewcommand{\arraystretch}{1.2}
\begin{tabular}{l|ccc|ccc|ccc}
\hline
\multirow{3}{*}{Dataset} & \multicolumn{3}{c|}{Llama3-8b-inst} & \multicolumn{3}{c|}{Qwen2.5-7b-inst}& \multicolumn{3}{c}{Qwen2.5 -1.5b-inst}\\ 
\cmidrule(lr){2-4} \cmidrule(lr){5-7} \cmidrule(lr){8-10}
& Raw & Refined & Refined & Raw & Refined & Refined & Raw & Refined & Refined \\
& & w/o ori & w ori & & w/o ori & w ori & & w/o ori & w ori \\ \hline
HWU64& 88.10& 87.17 {\scriptsize\color{blue}(-0.93)} & \textbf{89.03} {\scriptsize\color{red}(+0.93)}& 87.08& 85.97 {\scriptsize\color{blue}(-1.11)} & \textbf{88.38} {\scriptsize\color{red}(+1.30)}& 78.90& 79.27 {\scriptsize\color{red}(+0.37)} & \textbf{80.76} {\scriptsize\color{red}(+1.86)} \\
BANKING77& 85.88& \textbf{87.27} {\scriptsize\color{red}(+1.39)} & 87.95 {\scriptsize\color{red}(+2.07)}& 85.68& 86.33 {\scriptsize\color{red}(+0.65)} & \textbf{87.30} {\scriptsize\color{red}(+1.62)}& 73.31& \textbf{79.12} {\scriptsize\color{red}(+5.81)} & 77.34 {\scriptsize\color{red}(+4.03)} \\
CLINC150 & 95.03& 93.73 {\scriptsize\color{blue}(-1.30)} & \textbf{95.51} {\scriptsize\color{red}(+0.48)}& 95.36& 93.73 {\scriptsize\color{blue}(-1.63)} & \textbf{95.58} {\scriptsize\color{red}(+0.22)}& 82.29& \textbf{86.38} {\scriptsize\color{red}(+4.09)} & 84.02 {\scriptsize\color{red}(+1.73)} \\ \hline
CUREKART & 89.76& \textbf{91.94} {\scriptsize\color{red}(+2.18)} & \textbf{91.94} {\scriptsize\color{red}(+2.18)}& 90.02& \textbf{90.40} {\scriptsize\color{red}(+0.38)} & \textbf{90.40} {\scriptsize\color{red}(+0.38)}& 82.78& \textbf{86.27} {\scriptsize\color{red}(+3.49)} & 84.10 {\scriptsize\color{red}(+1.32)} \\
POWERPLAY11& 70.87& 76.05 {\scriptsize\color{red}(+5.18)} & \textbf{76.10} {\scriptsize\color{red}(+5.23)}& 71.20& 74.43 {\scriptsize\color{red}(+3.23)} & \textbf{74.76} {\scriptsize\color{red}(+3.56)}& 60.84& \textbf{66.99} {\scriptsize\color{red}(+6.15)} & 63.10 {\scriptsize\color{red}(+2.26)} \\
SOFMATTRESS& 82.61& 83.40 {\scriptsize\color{red}(+0.79)} & \textbf{85.40} {\scriptsize\color{red}(+2.79)}& 83.79& 82.60 {\scriptsize\color{blue}(-1.19)} & \textbf{87.40} {\scriptsize\color{red}(+3.61)}& 73.12& 79.44 {\scriptsize\color{red}(+6.32)} & \textbf{80.63} {\scriptsize\color{red}(+7.51)} \\ \hline
\end{tabular}
\caption{Complete performance comparison including refinement without original name preservation (Refined w/o ori). Results show that while both refinement approaches generally improve over the baseline (Raw), preserving original names during refinement (Refined w ori) tends to yield better or comparable results.}
\label{tab:Semantic vs Generic Label Refinemen_results}
\end{table*}
While our main experiments focus on semantic-aware refinement that preserves domain context, we also explore an alternative approach using generic identifiers (w/o ori). In this setting, all original intent labels are first replaced with generic identifiers (e.g., Intent\_1, Intent\_2) before refinement. During refinement, the model generates new labels without any influence from the original intent names.

The effectiveness of each approach varies with model size. Larger models (Llama3-8b-inst. and Qwen2.5-7b-inst.) generally perform better with original name preservation (w ori), particularly in BANKING77 where domain-specific terminology provides valuable semantic context. However, smaller models (Qwen2.5-1.5b-inst.) often show improved performance with generic identifiers (w/o ori), suggesting that they may benefit from the simplified label space. This performance pattern indicates that the choice between preserving or abstracting intent names should consider the model's capacity to leverage domain-specific terminology.

Complete results comparing both approaches across all datasets and models are shown in Table~\ref{tab:Semantic vs Generic Label Refinemen_results}.


\section{Full Prompt Template}
\label{appendix:prompt_details}

\subsection{Re-naming Prompt Template}

\begin{table}[ht]
\centering
\scalebox{0.85}{\begin{tabular}{l}
\hline
\begin{tabular}[c]{@{}p{1.0\linewidth}@{}}\\
\textbf{Examples by intent:}\\
\textbf{Intent 1:} account\_not\_verified\\
- How to Verify my Account?\\
- Account Verification\\
- Need to Verify my account\\
\\
\textbf{Intent 2:} delete\_pan\_card\\
- Pan card remove\\
- Delete PAN card\\
- I want to delete my pan card\\
\\
\textbf{Intent 3:} bank\_verification\_details\\
...\\
\textbf{Intent 4:} pan\_verification\_failed\\
...\\
\textbf{Rules for intent mapping:}\\
1. If the current intent name accurately represents its examples, map it to itself\\
2. If the intent name needs improvement, create a new descriptive name that better represents the examples\\
3. For new names:\\
   - Use lowercase letters only\\
   - Use underscores between words\\
\\
\textbf{INTENT MAPPINGS:}\\
account\_not\_verified ->\\
delete\_pan\_card ->\\
...\\
pan\_verification\_failed ->\\
\end{tabular}  \\ \hline
\end{tabular}}
\caption{Re-naming prompt template.}
\label{tab:renaming_prompt}
\end{table}
The re-naming prompt takes groups of examples organized by their original intent labels and analyzes their semantic meaning. Based on this analysis, it either keeps the original label if it accurately represents the examples, or generates a new more descriptive label while maintaining consistent formatting rules. The prompt enforces lowercase letters and underscores in label names to ensure standardization.

\subsection{Classification Prompt Template}
\begin{table}[]
\centering
\scalebox{0.85}{\begin{tabular}{l}
\hline
\begin{tabular}[c]{@{}p{1.0\linewidth}@{}}\\
You are an AI assistant specialized in intent classification. Your task is to determine\\
the single most likely intent of a given query based on the examples provided.\\
Provide only the name of the most probable intent, without any additional text or explanation.\\
\\
...\\
\\
\textbf{Text:} "Details for Bank Account Verification"\\
\textbf{Intent:} bank\_verification\_details\\
\\
\textbf{Text:} "Getting error while verifying PAN Card"\\
\textbf{Intent:} pan\_verification\_failure\\
\\
\textbf{Text:} "My PAN card needs to be verified"\\
\textbf{Intent:} verify\_pan\_card\_details\\
\\
\textbf{Query:} "My PAN card bank account verification please"\\
\textbf{The top 1 most likely intent is:}\\
\end{tabular}  \\ \hline
\end{tabular}}
\caption{Classification prompt template.}
\label{tab:classification_prompt}
\end{table}
The classification prompt presents example pairs of text queries and their corresponding refined intents to establish the task context. It instructs the model to determine the most likely intent for a new query based solely on these examples. The prompt explicitly requires only the intent name as output, without any additional explanation or text, to ensure consistent and clean predictions.

\subsection{Chain-of-Thought Prompt Template}
\begin{table}[]
\centering
\scalebox{0.85}{\begin{tabular}{l}
\hline
\begin{tabular}[c]{@{}p{1.0\linewidth}@{}}\\
You are an AI assistant specialized in intent classification. Your task is to determine\\
the single most likely intent of a given query based on the examples provided.\\
For each query:\\
1. Analyze the key elements and meaning\\
2. Provide an explanation of your reasoning\\
3. Extract the most likely intent\\
\\
\textbf{Text:} "Details for Bank Account Verification"\\
\textbf{Intent:} bank\_verification\_details\\
\\
\textbf{Text:} "Getting error while verifying PAN Card"\\
\textbf{Intent:} pan\_verification\_failure\\
\\
\textbf{Text:} "My PAN card needs to be verified"\\
\textbf{Intent:} verify\_pan\_card\_details\\
\\
\textbf{Query:} "My PAN card bank account verification please"\\
Provide your explanation and intent:
\end{tabular}  \\ \hline
\end{tabular}}
\caption{Chain-of-Thought prompt template.}
\label{tab:cot_prompt}
\end{table}

The Chain-of-Thought prompt extends the basic classification prompt by requiring the model to explain its reasoning process. For each query, the model must analyze the key elements, provide reasoning, and then determine the intent. This structured approach aims to help the model make more informed decisions by breaking down the classification process into steps.

\section{Related Work}
\subsection{Dialogue Intent Classification}
Dialogue intent classification aims to identify users' intentions from natural language utterances. Traditional approaches relied on supervised learning with large labeled datasets ~\cite{chen2019bertjointintentclassification, larson-etal-2019-evaluation}. The advent of large language models (LLMs) transforms this landscape, enabling effective few-shot learning approaches where only limited labeled data is available \cite{brown2020languagemodelsfewshotlearners}. This shift has been particularly significant as LLMs demonstrate strong few-shot capabilities through in-context learning, reducing the need for extensive labeled datasets \cite{liang-etal-2023-breaking,parikh2023exploringzerofewshottechniques,chen2024retrievalstyleincontextlearningfewshot}.

\subsection{Retrieval-based In-Context Learning}
Retrieval-based approaches have emerged as a powerful paradigm for improving few-shot learning performance. Key developments in this area include:
~\cite{milios2023incontextlearningtextclassification} propose effective retrieval strategies for in-context learning with many labels, demonstrating significant performance improvements through careful example selection. However, while this approach successfully retrieves semantically similar examples for classification, it introduces new challenges when dealing with intent labels that have high semantic overlap.~\cite{lu2024mitigatingboundaryambiguityinherent} This ambiguity between similar intents creates unnecessary complexity in the classification task, particularly when multiple intents share similar contextual meanings but require different downstream processing. Our work addresses this challenge by introducing a dynamic label refinement approach that helps distinguish between semantically similar intents while maintaining the benefits of retrieval-based example selection.

\section{Case Study}
\label{sec:case_study}

\subsection{Label Refinement Pattern Analysis}
Our analysis revealed interesting patterns in how the model refines intent labels, particularly highlighting some suboptimal refinement behaviors:

\subsubsection{Verbatim Query-to-Intent Conversion}
We observed cases where the model simply converted user queries directly into intent labels:

\begin{itemize}
    \item Original\_text: ``I want to return my mattress''
    \item Refined\_intent: \texttt{i\_want\_to\_return\_my\_mattress}
\end{itemize}

This pattern indicates a potential limitation in our refinement approach where the model sometimes fails to abstract the core intent, instead creating overly specific labels that mirror the input text.

\subsubsection{Overly Descriptive Intent Labels}
Another pattern emerged where the model generated unnecessarily verbose intent labels:

\begin{itemize}
    \item Original\_intent: \texttt{size\_customization}
    \item Refined\_intent: \texttt{how\_can\_i\_order\_a\_custom\_sized\_}\\ \texttt{mattress}
\end{itemize}
These findings highlight the need for more sophisticated label refinement strategies that maintain a balance between descriptiveness and practical utility.

\section{Performance comparison across different model combinations.}
\label{sec:combinations}
\begin{table}[h]
\centering
\adjustbox{width=0.48\textwidth}{ 
\setlength{\tabcolsep}{3pt}
\renewcommand{\arraystretch}{1.1}
\begin{tabular}{l||cc|cc}
\hline
\multirow{2}{*}{Dataset} & \multicolumn{2}{c|}{L3-8B} & \multicolumn{2}{c}{Q2.5-1.5B} \\ 
\cline{2-5}
& L3-8B & Q2.5-1.5B & Q2.5-1.5B & L3-8B \\ 
\hline
HWU64 & \textbf{89.03} & 80.95 & 80.76 & \textbf{88.48} \\
BANKING77 & \textbf{87.95} & 80.09 & 77.34 & \textbf{88.47} \\
CLINC150 & \textbf{95.51} & 86.82 & 84.02 & \textbf{94.79} \\
CUREKART& \textbf{91.94} & 84.96 & 84.10 & \textbf{91.07} \\
POWERPLAY11& \textbf{76.10} & 65.69 & 63.10 & \textbf{74.11} \\
SOFMATTRESS & \textbf{85.40} & 79.44 & 80.63 & \textbf{85.38} \\ 
\hline
\end{tabular}%
}
\caption{Performance comparison across different model combinations. The top row shows the model used for re-naming, while the models listed in the second row are used for answering (classification).}

\label{tab:model-combinations_appen}
\end{table}

Based on Table~\ref{tab:model-combinations}, we conduct additional experiments employing two separate large language models (LLMs) for intent segmentation and classification tasks in order to verify the effectiveness of label segmentation for each model. In particular, for the two models used in the experiment, we report the accuracy of both the model that generates the intents and the model that generates the responses. Consistent with the results shown in Table~\ref{tab:model-combinations}, we find that using a larger model for response generation is effective. Furthermore, we observe that label re-naming, even when performed using labels derived from a smaller model, still yields strong performance.

\end{document}